\documentclass[a4paper,table]{article}

\usepackage{INTERSPEECH2022}
\usepackage{cite}
\usepackage{xcolor}
\usepackage{highlight}
\usepackage{xspace}
\usepackage{hyperref}
\usepackage{diagbox}
\usepackage{multirow}

\title{AVATAR: Unconstrained Audiovisual Speech Recognition}
\name{Valentin Gabeur$^{1,2}$*, Paul Hongsuck Seo$^{2}$*, Arsha Nagrani$^2$*, \\
Chen Sun$^2$, Karteek Alahari$^1$, Cordelia Schmid$^2$
\thanks{\hspace{-12pt}* These authors contributed equally to this work. \\ $^\dagger$ Univ.\ Grenoble Alpes, CNRS, Grenoble INP, LJK, 38000 Grenoble, France.}
}
\address{
  $^1$Inria$^\dagger$ \hspace{1.3cm}
  $^2$Google Research}
\email{}

\makeatletter
\DeclareRobustCommand\onedot{\futurelet\@let@token\@onedot}
\def\@onedot{\ifx\@let@token.\else.\null\fi\xspace}

\def\eg{\emph{e.g}\onedot}

\makeatother

\begin{document}

\maketitle
 
\begin{abstract}

Audio-visual automatic speech recognition (AV-ASR) is an extension of ASR that incorporates visual cues, often from the movements of a speaker's mouth. Unlike works that simply focus on the lip motion, we investigate the contribution of entire visual frames (visual actions, objects, background etc.). This is particularly useful for unconstrained videos, where the speaker is not necessarily visible. To solve this task, we propose a new sequence-to-sequence AudioVisual ASR TrAnsformeR (AVATAR) which is trained end-to-end from spectrograms and full-frame RGB. To prevent the audio stream from dominating training, we propose different word-masking strategies, thereby encouraging our model to pay attention to the visual stream. We demonstrate the contribution of the visual modality on the How2 AV-ASR benchmark, especially in the presence of simulated noise, and show that our model outperforms all other prior work by a large margin. 
Finally, we also create a new, \textit{real-world} test bed for AV-ASR called VisSpeech, which demonstrates the contribution of the visual modality under challenging audio conditions.

\end{abstract}

\section{Introduction}
Automatic Speech Recognition (ASR) is often applied to edited or streamed media (for example, TV, online videos, video conferencing), where the input signal consists of both an audio and a visual stream. For these applications, the visual stream can provide strong cues for improving ASR, particularly in cases where the audio is degraded or corrupted. This has been largely exploited by AV-ASR works which focus on lip motion~\cite{Noda2014AudiovisualSR,Tamura2015AudiovisualSR,Chung2017LipRS,Afouras2018DeepAS,Petridis2018EndtoEndAS,Makino2019RecurrentNN,Ma2021EndToEndAS,Serdyuk2021AudioVisualSR} (using video crops centered around the speaker’s mouth). While lip motion is a strong signal in videos centered on the speaker, it may be less useful in some online videos (those with egocentric viewpoints, face coverings, poor video quality, speaker at a distance etc.). A more recent and less-explored direction is the contribution of additional visual context, for example, the hand movements of a speaker, the presence of certain objects that are being described or even the background location~\cite{Miao2016OpenDomainAS}.

In this paper, we focus on the latter case. 
The main existing benchmark for this task is the How2 dataset, which consists of instructional videos where the ground truth is obtained from user uploaded transcripts~\cite{sanabria2018how2}. While extremely valuable, the How2 dataset was created by keeping the audio samples that were most aligned to user-uploaded transcripts. This was done using an automatic alignment tool, and biases the dataset towards `clean' audio samples. We posit that in this case, a model trained on clean audio would never be incentivised to learn from the visual modality, as all the information for ASR is present in the audio stream. This has led to a number of AV-ASR works to doubt whether the visual modality is useful at all in a clean audio context, or if it is simply used as a regularizer~\cite{Caglayan2019VAT,Srinivasan2019AnalyzingUO,Ghorbani2021LLD}.

To resolve this issue, we explore masking strategies that degrade the audio samples during training, and then evaluate our model under noisy audio conditions. By dropping out key words in the audio signal, we incentivise our model to pay attention to the visual stream. 
Our model is based on a Seq2Seq encoder-decoder architecture. Unlike previous works that use full-frame pre-extracted visual features~\cite{Miao2016OpenDomainAS,Gupta2017VisualFeatures,Moriya2018LSTMLM,Palaskar2018EndtoendMS,sanabria2018how2,Caglayan2019VAT,paraskevopoulos2020multires,Srinivasan2020LookingEL,Ghorbani2021LLD}, our encoder performs audio-visual fusion early, and is trained directly from pixels and spectrograms. We show that large performance gains can also be achieved by pretraining our model on the large HowTo100M~\cite{miech19howto100m} dataset (not to be confused with How2).

Given the clean audio on the How2 test set, we also simulate noise at test time~\cite{Ghorbani2021LLD}. Unlike~\cite{Ghorbani2021LLD} which explicitly drops `visual words', we simulate noise in a more objective way, by randomly dropping audio segments, or adding external environmental sounds from the AudioSet dataset~\cite{Gemmeke2017AudioSet}. We show that under these conditions, our model with audio-visual inputs consistently outperforms an audio only model for the task of AV-ASR. 
In addition, we also create a small, \textit{real-world} test set with naturally occurring noisy audio.  This dataset is created by filtering out samples where automatic ASR gets perfect results, and hence is a challenging test bed. On this dataset, we show that the visual modality makes a significant contribution to performance.

In this work we make the following contributions: (i) We propose a new encoder-decoder AudioVisual ASR TrAnsformeR (AVATAR) which is trained end-to-end from spectrograms and full-frame RGB. Our encoder fuses audio and visual inputs and is trained jointly with the decoder;
(ii) To prevent the audio stream from dominating training, we propose and compare a number of masking strategies during training, thereby encouraging our model to pay attention to the visual stream; (iii) Our model achieves state-of-the-art performance on the How2 AV-ASR benchmark, with visual information improving performance under various simulated noise conditions; and finally (iv) we create a new, challenging \textit{real-world} test bed for AV-ASR called VisSpeech. The dataset is created using a combination of automatic techniques and manual annotation and unlike other works, allows us to demonstrate the performance of our method under realistic noise conditions. 
We have released this dataset publicly to the research community at \url{https://gabeur.github.io/avatar-visspeech}.

\section{Related Work}
\noindent\textbf{Audio-visual speech recognition.}
CTC~\cite{Graves2006ConnectionistTC} and Seq2Seq~\cite{Sutskever2014SequenceTS,Cho2014LearningPR} are the two most popular losses for performing ASR. In the context of AV-ASR focused on lip motion, they have been compared~\cite{Afouras2018DeepAS} and combined~\cite{Ma2021EndToEndAS}.
While early approaches~\cite{Noda2014AudiovisualSR,Tamura2015AudiovisualSR} use pre-extracted lip visual features, recent works~\cite{Chung2017LipRS,Afouras2018DeepAS,Petridis2018EndtoEndAS,Makino2019RecurrentNN,Ma2021EndToEndAS,Serdyuk2021AudioVisualSR} adopt an end-to-end approach by directly processing the pixels of the speaker's lips.
In contrast, the contribution of full frames for AV-ASR beyond the speaker's mouth movements has only been studied through pre-extracted visual context features: either action features~\cite{sanabria2018how2,Caglayan2019VAT,Ghorbani2021LLD,paraskevopoulos2020multires}, place features~\cite{Miao2016OpenDomainAS,Gupta2017VisualFeatures,Palaskar2018EndtoendMS,Caglayan2019VAT,Srinivasan2020LookingEL} or object features~\cite{Miao2016OpenDomainAS,Gupta2017VisualFeatures,Moriya2018LSTMLM,Palaskar2018EndtoendMS,Caglayan2019VAT,Srinivasan2020LookingEL}. Unlike these works, we train directly from full frame pixels for AV-ASR.

\noindent\textbf{Full-frame AV-ASR datasets.}
The How2 dataset~\cite{sanabria2018how2}, built from instructional videos, is the main benchmark for the full-frame AV-ASR task. Prior to this benchmark, full-frame AV-ASR works~\cite{Miao2016OpenDomainAS,Gupta2017VisualFeatures,Palaskar2018EndtoendMS} have also evaluated on instructional videos datasets but those have not been released. Audio-captioned image datasets~\cite{Srinivasan2019AnalyzingUO,Srinivasan2020LookingEL} have also been used for AV-ASR.
User-uploaded transcripts are the main source of ground truth for large scale AV-ASR video datasets~\cite{sanabria2018how2,Makino2019RecurrentNN,Serdyuk2021AudioVisualSR}. As these are often misaligned or inaccurate, transcripts are typically first audio-aligned and then automatically filtered~\cite{Liao2013Island}.

\noindent\textbf{Audio signal degradation for AV-ASR evaluation.}
In the case of lip motion for AV-ASR, several works have experimented with adding `babble noise'~\cite{Afouras2018DeepAS,Makino2019RecurrentNN} or extra speech tracks~\cite{Makino2019RecurrentNN,Serdyuk2021AudioVisualSR}. Additive Gaussian noise has also been used in~\cite{Serdyuk2021AudioVisualSR}.
In the case of full-frame AV-ASR, it is common to completely mask some segments of the audio signal that correspond to visual words. Srinivasan et al.~\cite{Srinivasan2020LookingEL} mask the audio segments corresponding to nouns and places. Ghorbani et al.~\cite{Ghorbani2021LLD} compute the similarity between words and visual frames and mask only the visual words.
We instead attempt to mimic more realistic settings by simulating audio degradations on How2 and releasing our `in the wild' benchmark VisSpeech. In order to prevent audio from dominating the task, we extend audio degradation to the training phase by randomly masking input words at training time.

\section{Method}

\subsection{Model Architecture} 
\begin{figure}[t]
  \centering
  \includegraphics[width=\linewidth]{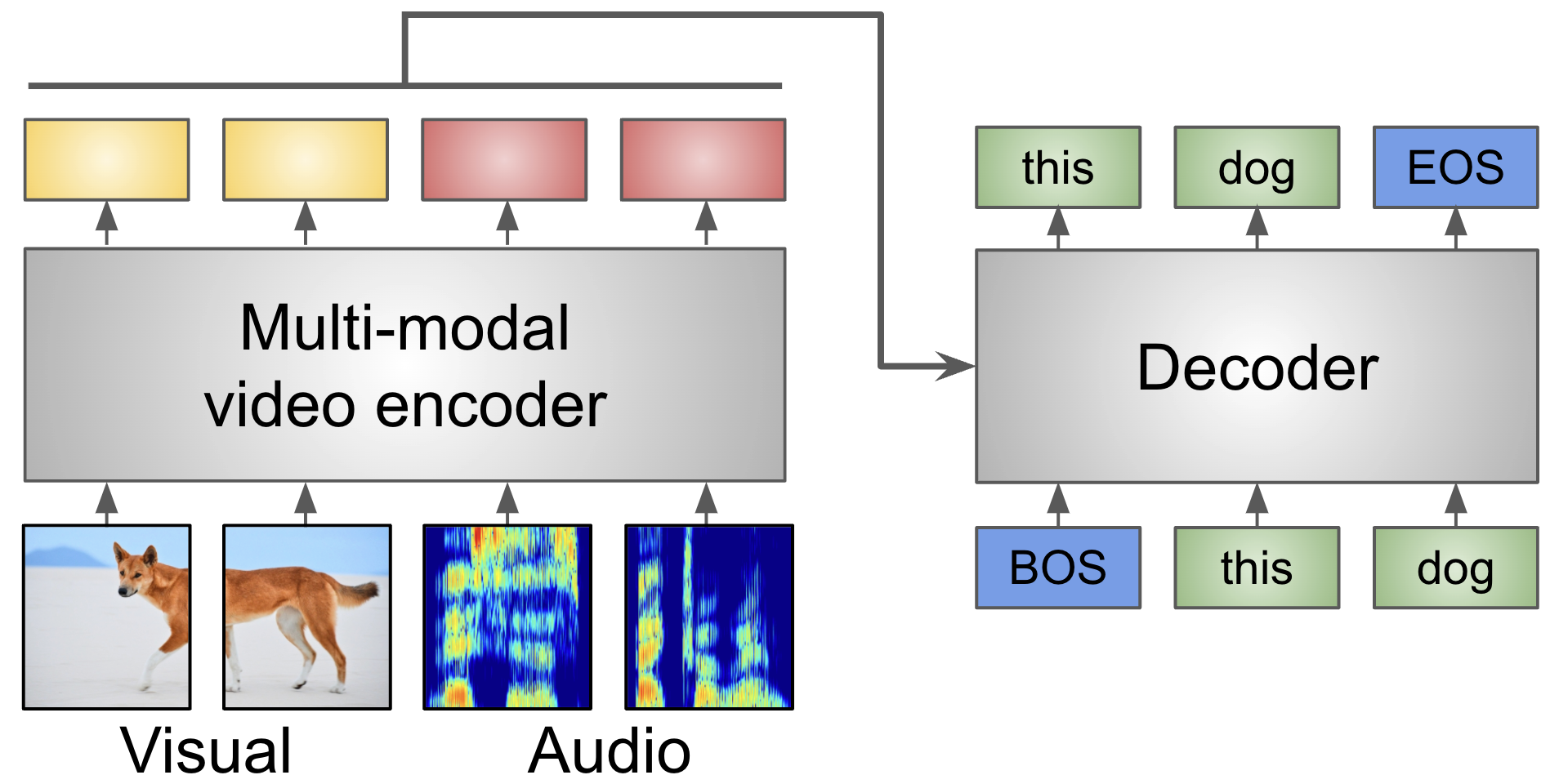}
  \caption{\textbf{AVATAR:} We propose a Seq2Seq architecture for audio-visual speech recognition. Our model is trained end-to-end from RGB pixels and spectrograms.}
  \label{fig:architecture_avatar}
\end{figure}

In this section we provide an overview of our audiovisual ASR model called AVATAR (Figure~\ref{fig:architecture_avatar}). 
Our model consists of a multimodal encoder to encode both RGB frames and audio spectrograms, and a transformer decoder which produces the natural language speech recognition output. Unlike previous AV-ASR works, we do not use frozen visual features, but have a single multimodal encoder that allows early multimodal fusion~\cite{nagrani2021attention}.\\
\noindent\textbf{Audio Inputs.} Our model operates on 25 second audio inputs. We follow common practice and extract 80-dimensional filter bank features from the 16kHz raw speech signal using a Hamming window of 25ms and a stride of 10ms, giving us $80\times2500$ size spectrograms for 25 seconds of audio. We then extract $16\times16$ non overlapping patches, giving us a total of $5\times156 = 780$ input tokens for audio. \\
\noindent\textbf{RGB Inputs.} 
We randomly extract 2 frames at 2.5 fps from each input video clip, which are then converted into tokens by extracting $16\times16\times2$ tubelets resulting in a total of $14\times14 = 146$ input tokens (the image resolution is $224\times224$).
This is based on our observations that visual signals are highly redundant for most videos and can therefore be efficiently captured from few frame samples. \\
\noindent\textbf{Audiovisual Encoder.}
We adopt the recently proposed MBT architecture~\cite{nagrani2021attention}, which is a transformer based multimodal encoder. 
Given both sets of audio and RGB tokens, MBT first adds positional encodings to each token, and append a CLS token to each set.
The sets are then fed to the MBT encoder.
MBT relies on bottleneck tokens to model the cross-modal interactions. 
Here we use the best parameters from~\cite{nagrani2021attention} (4 bottleneck tokens and bottleneck fusion starting at layer 8).
We use public ViT-Base~\cite{50650} weights (ViT-B, $d_{model}=768$ $L=12$, $N_H=12$, $d=3072$)\footnote{$d_{model}$ is the embedding dimension, $L$ is the number of transformer layers, $N_H$ is the number of self-attention heads with hidden dimension $d$.} pretrained on ImageNet-21K~\cite{deng2009imagenet} for initialization.\\
\textbf{Decoder.} All hidden units from the encoder are then passed to an auto-regressive transformer decoder~\cite{vaswani2017transformer} consisting of 8 layers and 4 attention heads. 

\subsection{Training Strategies} 
In this section we describe our training loss and strategy for AVATAR. 
Our model is first pretrained on a large dataset with transcripts obtained using an \textit{audio-only} ASR API~\cite{miech19howto100m}. To entice the model to pay attention to the visual modality, we introduce a word masking strategy, which is described below.\\
\textbf{End to end training.} The model is trained end-to-end using a cross-entropy loss on each decoded token. \\
\textbf{Word Masking.} To prevent our model from ignoring the visual modality, we introduce word masking techniques during training.
We randomly sample target words and mask out the input audio signals that correspond to those words using pre-extracted alignments between words and the input signal.
We obtain the alignment either from ASR results for pretraining or by using an off-the-shelf forced-alignment tool for finetuning.
For selecting target words to mask, we experiment with two strategies: random and content word masking.
The former strategy selects target words randomly whereas the latter chooses the targets among non-stop (henceforth known as content) words.
For random word masking, we randomly mask out 10\% of the words.
For content word masking, there are fewer candidate words to be masked so content words are masked at a higher rate to match the 10\% overall masking on the entire dataset.

\section{VisSpeech Dataset}
In this section we describe our new AV-ASR benchmark called VisSpeech. Our dataset is a subset of the publicly released HowTo100M dataset~\cite{miech19howto100m},
and is curated using a combination of automatic filtering stages and manual verification. \\
\noindent\textbf{Dataset creation pipeline:}\\
Our dataset creation pipeline is driven by two objectives: (1)~We want to find challenging audio conditions in which regular audio-based ASR fails. For this we seek videos where there is a large word error rate between automatic ASR and user-generated transcripts. (2)~We are also interested in video segments where there is \textit{high audio-visual correspondence}, in order to create a suitable multimodal test set for AV-ASR. Our pipeline consists of the following steps: \\
\textbf{Step 1: Obtain videos with user uploaded transcripts.}
We first search for HowTo100M videos that have both manually uploaded transcripts and automatic ASR. We then align the transcripts using the Levenshtein algorithm in order to compute the global word error rate (WER) between the two.
Videos with a global WER greater than 100\% are removed, as we find this helps filter out completely wrong user uploaded transcripts or ASR. 
This gives us 85K videos. 
\\ 
\textbf{Step 2: ASR vs user transcripts.}
Videos are split into segments, at silences detected using an open-source VAD model~\cite{Silero_VAD}.
Using the user transcripts and the ASR from each segment,
we filter out segments with WER greater than 50\% between the two transcripts to remove samples with significantly low quality user transcripts or ASR. 
This is similar to the rationale for the filtering in Step 1, which is performed at a video level, but now we perform it at a segment level.
Next and most importantly we remove ``easy" samples, namely those with low WER of less than 20\% for non-stop words (too clean) and samples with less than 9 words (too short). 
This leaves us with 773K sentences from 7.5K videos.
\\ 
\textbf{Step 3: Visual-Text similarity.}
To measure visual-text similarity, we run a video-text similarity model \cite{nagrani2022learning} trained on the Howto100M dataset to get similarity scores between each video and sentence pair. This helps highlight challenging samples where the visual modality can compensate for corrupted audio.  \\ 
\textbf{Step 4: Manual annotation.} Finally, we manually check the highest similarity segments and correct the user-uploaded ASR if necessary. 

VisSpeech consists of 508 segments from 495 unique videos. The average transcript length is 12.2 words and average segment duration is 4.3 seconds. By filtering for segments where ASR fails, we find that our dataset is truly a challenging test bed `in the wild', with the audio containing background chatter, laughter, music and other environmental sounds. During the manual verification phase, we also noticed that many examples contain speech spoken with challenging English accents from various regions all over the world.

\section{Experiments}

\begin{table*}[h]
  \caption{Audiovisual ASR vs Audio only models under various \textit{evaluation} noise conditions (Clean, Burst, Environment and Mixed) and with different \textit{training} masking strategies (Random and Content). Percentage Word Error Rate (\%WER) is reported on the How2 test set. \textbf{A:} Audio-only. \textbf{A+V:} Audiovisual. \textbf{Rel. $\Delta$:} Relative improvement of A+V over A.}
  \label{tab:noise}
  \centering
  \begin{tabular}{lc@{\hskip 3mm}c@{\hskip 3mm}c@{\hskip 3mm}c@{\hskip 3mm}c@{\hskip 3mm}c@{\hskip 3mm}c@{\hskip 3mm}c@{\hskip 3mm}c@{\hskip 3mm}c@{\hskip 3mm}c@{\hskip 3mm}c}
 \toprule
 \multirow{2}{*}{\backslashbox{Training}{Eval Noise}}& \multicolumn{3}{c}{Clean} & \multicolumn{3}{c}{Burst Loss} & \multicolumn{3}{c}{Environment Noise} & \multicolumn{3}{c}{Mixed Noise} \\
 & \textbf{A} & \textbf{A+V} & \textbf{Rel. $\Delta$} & \textbf{A} & \textbf{A+V} &\textbf{Rel. $\Delta$} & \textbf{A} & \textbf{A+V} &\textbf{Rel. $\Delta$} & \textbf{A} & \textbf{A+V} &\textbf{Rel. $\Delta$}\\
    \midrule
    No Pretraining & 15.72 & 15.62 & \gradient{0.64} & 29.59 & 28.69 & \gradient{3.05} & 50.79 & 47.70 & \gradient{6.08} & 60.51 & 57.49 & \gradient{5.0} \\
    Vanilla & 9.75 & 9.79 & \gradient{-0.33} & 21.97 & 21.71 & \gradient{1.19} & 25.97 & 25.55 & \gradient{1.61} & 39.13 & 38.96 & \gradient{0.42}\\
    Random Word Masking & 9.19 & 9.11 & \gradient{0.93} & 15.60 & 15.28 & \gradient{2.05} & 23.39 & 22.35 & \gradient{4.45} & 32.43 & 30.64 & \gradient{5.50} \\
    Content Word Masking & 9.58 & 9.25 & \gradient{3.48} & 17.26 & 16.92 & \gradient{1.98} & 23.77 & 22.67 & \gradient{4.65} & 33.83 & 32.26 & \gradient{4.53} \\
    \bottomrule
  \end{tabular}
\end{table*}
\begin{figure*}[h!]
  \centering
  \includegraphics[trim={0cm 0cm 0cm 0}, width=\linewidth]{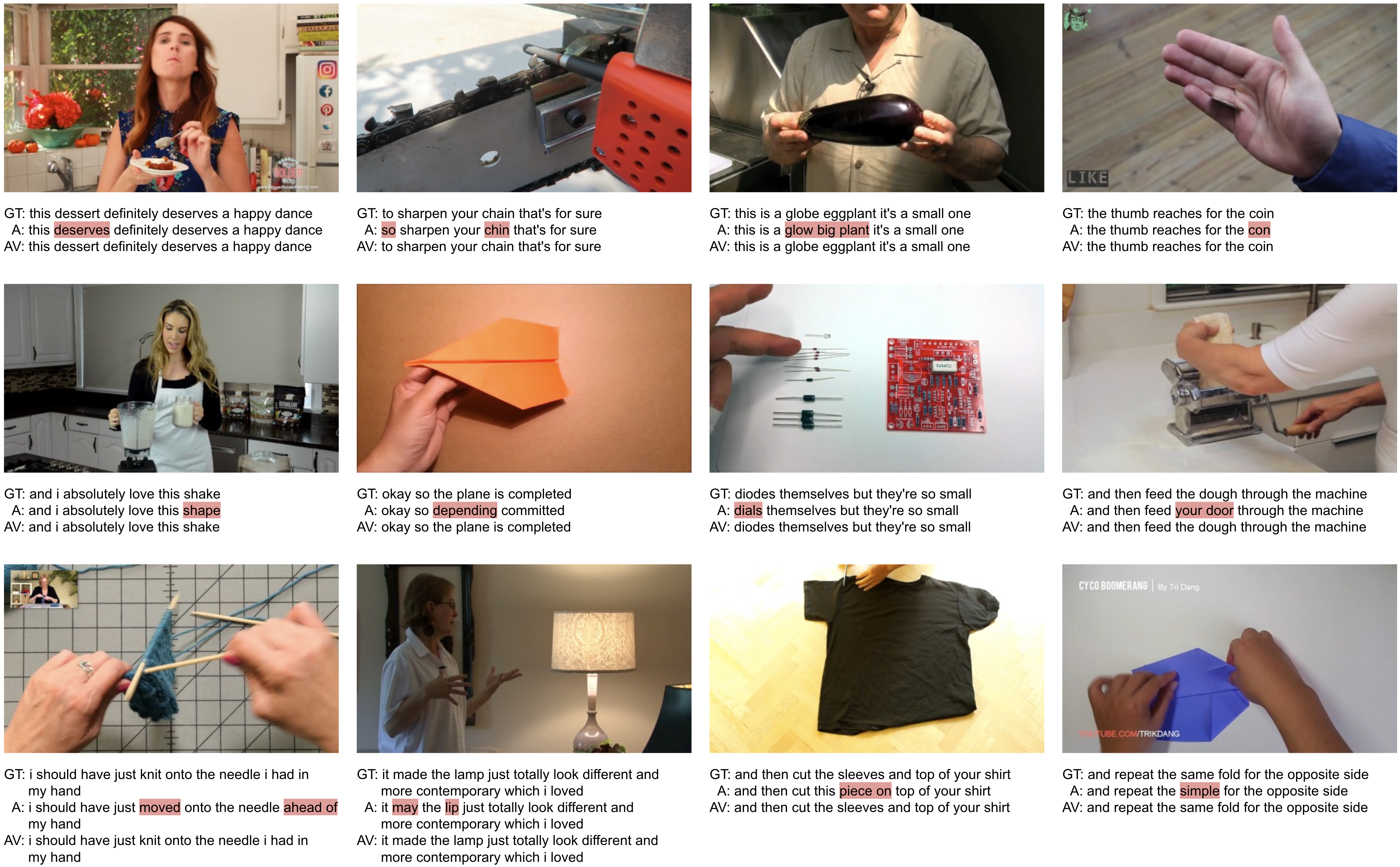}
  \caption{\textbf{Qualitative results on the VisSpeech dataset.} We show the ground truth (GT), and predictions from our audio only (A) and audio-visual model (A+V). Note how the visual context helps with objects (`chain', `eggplant', `coin', `dough'), as well as actions (`knit', `fold') 
  which may be ambiguous from the audio stream alone. Errors in the predictions compared to the GT are highlighted in red. 
  }
  \label{fig:qualitative_visspeech}
\end{figure*}

In this section we first describe the datasets, metrics and implementation details for training (Section~\ref{sec:data}). We then describe the simulated noise we use for evaluating our models on the How2 dataset (Section~\ref{sec:noise}). Finally we discuss the results of our model under different training strategies on both the How2 and our newly introduced VisSpeech dataset (Section~\ref{sec:results}). 

\subsection{Datasets, Metrics, Implementation Details}\label{sec:data}

\noindent\textbf{HowTo100M~\cite{miech19howto100m}} consists of more than 1 million instructional videos associated with their automatically-extracted speech transcriptions. We only use this dataset for pre-training our model. Note that here we are not training with perfect ground truth, but using the ASR outputs of an existing model. We remove videos present in the validation and test sets of VisSpeech and How2 datasets (described next).

\noindent\textbf{How2~\cite{sanabria2018how2}} is an instructional video dataset created for multi-modal language understanding. We use the 300 hour version. The videos are segmented into short clips (avg 5.8s), each accompanied by their user-uploaded transcript (avg 20 words). The dataset is split between a training (184,949 clips), validation (2,022 clips) and test (2,305 clips).

\noindent\textbf{Metrics.} We evaluate our models using Word Error Rate (WER). For each sentence, dynamic programming is used to align the predicted words to the ground truth. The number of word errors (deletions, substitutions and insertions) is then computed across the whole test dataset and divided by the number of ground truth words to obtain the WER. 

\noindent\textbf{Training Implementation details.}
All models are trained end-to-end unless otherwise specified.
We use a batch size of 1,536 and 256 for pretraining and finetuning respectively.
We adopt a wordpiece tokenizer~\cite{wu2016google} pretrained for BERT and decode using a beam search with a beam size of 4 and a brevity penalty of 0.6.
We use SpecAugment with parameters adopted from \cite{Park2019SpecAugmentAS}.
We augment the visual frames using random cropping and color jittering. 
We use a momentum optimizer. We pretrain our model for 1M iterations with an initial learning rate of 2. The learning rate is warmed up for 1K iterations and then linearly decayed to 0.
We initialize both visual and audio streams of the MBT encoder with the public ViT~\cite{50650} weights pretrained on ImageNet.
For finetuning, we train for 40K iterations without warmup.

\subsection{Simulated Noise for Evaluation}\label{sec:noise} 
The ground-truth transcripts in How2 are collected by performing forced-alignment on the user uploaded transcripts, and filtering out examples with a low confidence score.
Due to such a filtering process, the audio signals in How2 are inherently clean, and consequently the task is largely audio dominant. To overcome this limitation, we evaluate our models with three types of simulated audio noise: burst packet loss, environment noise and mixed noise.
For the burst packet loss, we randomly drop two chunks of the input audio signal where the length of each chunk is uniformly sampled from $(0, 0.1]$ times the video duration.
To simulate environment noise, we add audio noise randomly sampled from the `noise' and `environmental' classes in the AudioSet dataset~\cite{Gemmeke2017AudioSet}. 
Finally, we also evaluate a combination of the two (`mixed noise').
Note that we train a single model and evaluate it under different noise configurations.

\subsection{Results}\label{sec:results}

\noindent\textbf{Effects of Training Strategies.}
Table~\ref{tab:noise} compares audio-only (A) and audiovisual (A+V) models trained with different training strategies and evaluation noise settings on How2. 
In addition to clean audio, we report results in three degraded audio scenarios, burst loss, environment noise and mixed noise. 
We first note that without pretraining, our model performs well on the clean eval, but performance degrades significantly under simulated noise conditions. Adding the visual modality under these noise conditions helps performance across the board. 
Vanilla pretraining then improves performance significantly, however we note the gap between A and A+V also shrinks,  signifying the improvement is largely from better audio encoding. In this case, the A+V model has no incentive to look at visual inputs, as the task under clean conditions is dominated by audio. In addition, the pretraining is performed on HowTo100M where the transcripts are automatically generated from the audio alone, and so our model is able to solve the task without any visual information. We find the word masking strategies to be extremely effective to mitigate this. The overall performance of both A and A+V are improved and notably, the improvements of A+V are larger. We show that with these masking schemes, adding visual inputs helps even for clean audio, with the performance improving under environment and mixed noise. 
Note that across the board, adding the visual modality improves performance.

\begin{table}[t]
  \caption{\textbf{Comparison to the state-of-the-art on How2.} Our model outperforms all previous works when trained from scratch, and pretraining provides a significant boost. We report the best audio-visual numbers for all works.}
  \label{tab:how2_sota}
  \centering
  \begin{tabular}{ lc}
    \toprule
    \textbf{Model} & 
    \textbf{\%WER}\\
    \midrule
    BAS~\cite{sanabria2018how2} & 
     18.0 \\
    VAT~\cite{Caglayan2019VAT} &  
    18.0 \\
    MultiRes~\cite{paraskevopoulos2020multires} & 
    20.5 \\
    LLD~\cite{Ghorbani2021LLD} & 
    16.7 \\
       \midrule
    AVATAR (scratch) &  
    15.6 \\
    AVATAR (pretrained) & 
\textbf{9.1} \\
    \bottomrule
  \end{tabular}
\end{table}

\renewcommand*{\minval}{0.0}
\renewcommand*{\maxval}{11.30}
\begin{table}[t]
  \caption{WERs of AVATAR on our newly introduced test set VisSpeech consisting of real-world noise. The models are trained on automatic ASR from HowTo100M, and finetuned on How2. Note here we do not add any artificial audio degradation at all.}
  \label{tab:vispeech_results}
  \centering
  \begin{tabular}{ lccc}
    \toprule
    \textbf{Training Strategy} & 
    \textbf{A} & 
    \textbf{A+V} & \textbf{Rel. $\Delta$} \\
    \midrule
    No pretraining & 44.57 & 43.41 & \gradient{2.61} \\
    Vanilla & 12.69 & 11.91 & \gradient{6.11} \\ 
    Random Word Masking & 12.35 & 11.86 & \gradient{3.93} \\ 
    Content Word Masking & 12.72 & 11.28 & \gradient{11.30} \\
    \bottomrule
  \end{tabular}
\end{table}

\noindent\textbf{Comparison to the state-of-the-art.}
Table~\ref{tab:how2_sota} compares AVATAR with existing state-of-the-art methods on How2. Our model trained from scratch already outperforms all existing methods and serves as a strong baseline.
Our best model, which is pretrained on HowTo100M, with the random word masking technique, brings a further boost reducing the error rate by over 45\% relatively compared to the existing state-of-the-art method.

\noindent\textbf{Evaluation on VisSpeech.}
We evaluate our AVATAR model trained with different strategies on VisSpeech with real-world noise (Table~\ref{tab:vispeech_results}).
We finetune the pretrained models for 5K iterations on How2.
Our dataset effectively highlights the contribution of the visual modality without introducing any artificial noise.
Once again, both masking strategies help the audiovisual (A+V) model learn to utilize the visual modality better.
Content word masking improves the performance only when the visual modality is provided; providing some evidence that the A+V model uses visual inputs to correct errors on content words. To further tease apart the input of the visual modality, we compute the word error rate on content words \textit{only} and on stop words \textit{only}. This is because we hypothesize that visual modality should not be able to provide any useful information about stop words, and so most of the improvement should be on the content words. As expected, we find the errors on content words are reduced substantially more than those on stop words when the visual modality is incorporated with all training strategies (\eg, 1.18\% vs. 0.33\% absolute error rate drops with content word masking).
This also confirms the contribution of the visual modality.
Further evidence can be found from the qualitative examples provided in Figure~\ref{fig:qualitative_visspeech}, where it can be clearly seen that visual context helps with correcting ASR errors on objects as well as actions. 

\noindent\textbf{End-to-end training with early audiovisual fusion.}
Unlike previous works on full-frame AV-ASR, AVATAR is trained (i) entirely end-to-end, and (ii) with early audiovisual fusion in the MBT encoder.
To assess this effect, we test AVATAR with pre-extracted visual features as in~\cite{Ghorbani2021LLD}, using a model pretrained on HowTo100M with NCE loss \cite{nagrani2022learning}. 
We concatenate the audio features at the output of our MBT encoder with our pre-extracted visual features and provide them to the decoder.
Note that in this case the audio-visual fusion happens only through the decoder.
We train this model with random masking strategy and 
find that the end-to-end trained model outperforms the model with pre-extracted features with 9\% relative improvement in the mixed noise setting and similar trends are observed in all the other noise settings. \\
\noindent\textbf{Contribution of Visual Modality.}
Some works show that the visual modality is simply a regularizer~\cite{Caglayan2019VAT,Srinivasan2019AnalyzingUO,Ghorbani2021LLD}. As done by \cite{Srinivasan2019AnalyzingUO}, we further investigate whether the contribution of the visual modality is simply a regularizer by replacing the visual frames of test examples with those extracted from random validation videos.
Unlike \cite{Srinivasan2019AnalyzingUO}, we observe significant degradation of A+V models in all settings (\eg, 9.11\%$\rightarrow$9.53\% with random word masking and clean audio) as the models get distracted by the random visual inputs. Note that this performance (9.53\%) is worse than the audio-only model in this setting (9.19\%).
This suggests vision in our model is not simply a regularizer contrary to what was previously reported in~\cite{Caglayan2019VAT,Ghorbani2021LLD}.

\section{Conclusion}
In this work, we propose a novel encoder-decoder transformer architecture and training strategies based on word masking for AV-ASR.
We show that our method helps the model learn to use visual inputs better and outperform the state of the art.
Finally we also release VisSpeech, a new AV-ASR test benchmark, and demonstrate the effectiveness of our method under naturally occurring noise.

\noindent {\bf Acknowledgements.} This work was supported in part by the ANR
grant AVENUE (ANR-18-CE23-0011).

\bibliographystyle{IEEEtran}
\bibliography{main}

\end{document}